 \documentclass[final,5p,times,twocolumn,authoryear]{elsarticle}

\usepackage{amssymb}
\usepackage{lipsum}
\journal{Open}

\usepackage[acronym]{glossaries}

\newacronym{ml}{ML}{machine learning}
\newacronym{llm}{LLM}{large language model}
\newacronym{sota}{SotA}{state-of-the-art}
\newacronym{nlp}{NLP}{natural language processing}
\newacronym{ai}{AI}{artificial intelligence}
\newacronym{dl}{DL}{deep learning}
\newacronym{rl}{RL}{reinforcement learning}
\newacronym{gai}{GenAI}{Generative AI}
\newacronym{cot}{CoT}{chain-of-thought}
\newacronym{lms}{LMS}{learning management system}
\newacronym{llama}{LLaMA}{Large Language Model Meta AI}

\begin{document}

\begin{frontmatter}

%% Title, authors and addresses

%% use the tnoteref command within \title for footnotes;
%% use the tnotetext command for theassociated footnote;
%% use the fnref command within \author or \affiliation for footnotes;
%% use the fntext command for theassociated footnote;
%% use the corref command within \author for corresponding author footnotes;
%% use the cortext command for theassociated footnote;
%% use the ead command for the email address,
%% and the form \ead[url] for the home page:
%% \title{Title\tnoteref{label1}}
%% \tnotetext[label1]{}
%% \author{Name\corref{cor1}\fnref{label2}}
%% \ead{email address}
%% \ead[url]{home page}
%% \fntext[label2]{}
%% \cortext[cor1]{}
%% \affiliation{organization={},
%%            addressline={}, 
%%            city={},
%%            postcode={}, 
%%            state={},
%%            country={}}
%% \fntext[label3]{}

\title{Counterargument for Critical Thinking as Judged by \acrshort{ai} and Humans}

%% use optional labels to link authors explicitly to addresses:
%% \author[label1,label2]{}
%% \affiliation[label1]{organization={},
%%             addressline={},
%%             city={},
%%             postcode={},
%%             state={},
%%             country={}}
%%
%% \affiliation[label2]{organization={},
%%             addressline={},
%%             city={},
%%             postcode={},
%%             state={},
%%             country={}}

\author{Tosin Adewumi\textsuperscript{*}, Marcus Liwicki, Foteini Simistira Liwicki, Lama Alkhaled, Hamam Mokayed, Esra Sümer-Arpak\\ Machine Learning Group, EISLAB, \\Luleå University of Technology, Sweden. \\
firstname.lastname@ltu.se}

\begin{abstract}
%% Text of abstract
This intervention study investigates the use of counterarguments in writing for critical thinking by students in the context of \acrfull{gai}.
This is especially as risks of cheating and cognitive offloading exist with the use of \acrshort{gai}.
We presented 36 students in a particular university course with 4 carefully selected thesis statements  (from a set of popular debates) to write about anyone of them.
We used six established rubrics (focus, logic, content, style, correctness and reference) to conduct three human assessments (two student peer-reviews and one experienced teacher) per writeup on a 5-point Likert scale  for all the qualified samples (\textit{n}) of 35 submissions (after disqualifying one for irregularity).
Using the same rubrics and guidelines, we also assessed the submissions using six frontier \acrshort{llm}s as judges.
Our mixed-method design included qualitative open-ended feedback per assessment and quantitative methods.
The results reveal that (1) the students' self-written counterarguments to \acrshort{ai}-generated content contains logic, among other things, which is a key component of critical thinking, and (2) \acrshort{gai} can be successfully used at scale to assess students’ written work, based on clear rubrics, and these assessments generally align with human assessments as shown with Gwets AC2 inter-rater reliability values of 0.33 for all the models except one.

\end{abstract}
% Background (1–2 sentences), Research gap, Purpose

% Computers and Education: Artificial Intelligence - Journal name

%%Graphical abstract
%\begin{graphicalabstract}
%\includegraphics{grabs}
%\end{graphicalabstract}

%%Research highlights
%\begin{highlights}
%\item Research highlight 1
%\item Research highlight 2
%\end{highlights}

\begin{keyword}
%% keywords here, in the form: keyword \sep keyword, up to a maximum of 6 keywords
Counterargument \sep Argument-Based Learning \sep Critical Thinking \sep \acrshort{ai}

%% PACS codes here, in the form: \PACS code \sep code

%% MSC codes here, in the form: \MSC code \sep code
%% or \MSC[2008] code \sep code (2000 is the default)

\end{keyword}

\end{frontmatter}

%\tableofcontents

%% \linenumbers

%% main text

\section{Introduction}
\label{introduction}

Education is evolving rapidly, particularly its pedagogy \citep{zou2025navigating}, the engagement and evaluation of students \citep{ennis1985ennis,halpern2006halpern}, and the end-to-end planning of learning \citep{romiszowski2016designing}.
This is thanks to \acrfull{gai}, particularly \acrlong{llm}s (\acrshort{llm}s), which have many benefits \citep{11364207,mulaudzi2025lecturer}.
Recent studies, however, have shown that people who used ChatGPT or similar tools to write essays demonstrated less brain activity in areas linked with cognitive processing or reported negative impact to their skills \citep{kosmyna2025your,helal2025impact}.
This behaviour of shifting cognitive tasks to \acrshort{gai} is termed cognitive offloading and may lead to cognitive atrophy, where one's abilities become worse \citep{kosmyna2025your,gerlich2025ai}.
Given that \acrshort{gai} is most likely here to stay, it is essential and beneficial to find ways that students can engage with it such that it encourages critical thinking, thereby boosting their skills and learning.

Critical thinking is a purposeful, reflective meta-cognitive process that increases the possibility of forming a robust and logical conclusion to an argument or finding a solution to a problem \citep{facione1990delphi,ku2009assessing,lai2011critical,dwyer2014integrated}.
It emphasizes moving beyond passive information intake to examining evidence, identifying assumptions, comparing perspectives, and using logic to determine the strength of a position \citep{kuhn1991skills, sinfield2023design}.
Meanwhile, a counterargument is a logical argument that offers a different viewpoint to an existing one, thereby constituting a debate \citep{fulan2025corpus}.
While there are differences of opinion on what the components of critical thinking are \citep{dwyer2014integrated}, logic is widely seen as a fundamental ingredient \citep{adewumi2023procot,dwyer2014integrated}.

Assessing critical thinking through open-ended formats, such as with written counterarguments, is favored by many researchers because the format seeks to evaluate the logic in an existing content.
One example of an open-ended format is the Ennis–Weir Critical Thinking Essay Test (EWCTET) \citep{ennis1985ennis,lai2011critical}.
However, objectively assessing students' writing for critical thinking can be a challenging task for many reasons, including the differences of opinion on the details of critical thinking \citep{dwyer2014integrated}.
The Likert scale offers a possible solution and has been used by other researchers \citep{joshi2015likert,alkharusi2022descriptive,gerlich_ai_2025}.
%\cite{joshi2015likert} discuss the argument for 7 or 5-point Likert scale.

% The Assessment Problem Human scoring limitations (bias, cost, inconsistency) AI-based scoring emergence The alignment problem (AI vs human judgment)

In this work, our motivation is to answer \textbf{two research questions}: (1) Do counterarguments to \acrshort{ai}-generated arguments promote students' critical thinking in writing, in terms of logic and relevant rubrics? and (2) How similar (or dissimilar) are \acrshort{gai} systems in judging counterarguments compared to humans, given the same rubrics?
This is more so that the topic of how student-written counterarguments to \acrshort{ai}-generated arguments across diverse topics promote critical thinking in students is understudied in the literature, since argument-based learning appears under-utilized in pedagogy compared to traditional transmissive teaching methods \citep{gonzalez2026promoting}.
We used a mixed \textbf{research design} involving qualitative and quantitative methods, with a sample size (\textit{n}) of 35 submissions and the 5-point Likert scale for the rubrics.
We publicly release the data artifacts.\footnote{github.com/LTU-Machine-Learning/counterargument\_ai}
Our key contributions are: 
\begin{enumerate}
    \item We show that students' self-written counterarguments to \acrshort{ai}-generated content contains logic, among other things, which is a key component of critical thinking.
    
    \item We show that \acrshort{gai} can be successfully used to assess students' written work (particularly counterarguments) based on clear rubrics and its assessments generally align with human assessments, including expert review and students' peer review.    
\end{enumerate}

The rest of this paper is organised as follows.
In Section \ref{theoretical}, we describe the theoretical framework in the literature.
In Section \ref{method}, we fully describe the method of this study, including the research design, participants and the checks for \acrshort{ai}-generated counterarguments.
In Section \ref{results}, we present our findings using multiple charts.
In Section \ref{discussion}, we discuss the implications of this study in connection to theory.
Finally, in Section \ref{conclusion}, we provide concluding remarks.

\section{Theoretical Framework}
\label{theoretical}

We conduct a relatively thorough review of the literature on some of the theoretical underpinnings of critical thinking, counterarguments, and \acrshort{gai} assessment.

\subsection{Models of Critical Thinking}
\cite{bloom1956taxonomy}'s taxonomy of educational objectives is one of many models of thinking applicable to critical thinking because it contains the analysis of knowledge \citep{dwyer2014integrated}.
The model has influenced many others, including \cite{anderson2001taxonomy}'s revised taxonomy for learning, teaching, and assessing, \cite{duron2006critical}'s 5-step framework, and \cite{romiszowski2016designing}'s course design using a systems approach.

In the first layer of the model by \cite{bloom1956taxonomy} is \textit{knowledge}, which not only relates to the specifics and terminologies of a content but knowledge about the ways of dealing with those specifics.
In the second layer is \textit{comprehension}, involving explaining and summarizing learned information, while in the third layer is \textit{application}.
In the fourth layer is \textit{analysis} of elements and how they relate to one another.
In the fifth layer is \textit{synthesis}, which involves the production of a plan or new communication, while in the final, sixth, layer is \textit{evaluation}.
The Delphi panel (of 46 experts in critical thinking) agreed that analysis, evaluation and inference are core skills for critical thinking \citep{facione1990delphi} and they positively correlate \citep{dwyer2015promotion}.
The three skills form crucial components of the integrative critical thinking framework by \cite{dwyer2014integrated}.

\subsection{Counterargument as a Tool of Critical Thinking}

Beyond their structural role in argumentation, counterarguments function as a critical tool in argument-based learning (e.g. argumentative writing and debate), serving to strengthen a primary argument by acknowledging, analyzing, and refuting opposing viewpoints \citep{gonzalez2026promoting}.
Indeed, the definition of critical thinking is fully entrenched in the making of logical arguments (or counterarguments) \citep{dwyer2014integrated}.
%Instead of weakening a claim, addressing counterarguments proves that the writer has considered multiple sides of an issue, enhancing credibility and demonstrating academic rigor \citep{chi2014icap, nussbaum2007promoting}. 
In educational settings, exposure to contrasting viewpoints has been shown to stimulate conceptual change and promote integrative understanding, particularly when students are required to respond to those alternatives.
This is the case either with the concept of \textit{argue to learn} (i.e.  facilitating the learning of field knowledge) or \textit{learn to argue} (i.e. pedagogical tool for developing critical thinking skills) \citep{chi2014icap,
nussbaum2003argument,gonzalez2026promoting}.

In the context of \acrshort{gai}, emerging research suggests that critical interaction with LLM outputs, such as questioning, revising, or challenging generated text, can support higher-order thinking when learners remain cognitively active  \citep{kasneci2023chatgpt}. Rather than accepting AI-generated responses at face value, students who formulate counterarguments must evaluate the adequacy, relevance, and coherence of the presented reasoning. 
The use of counterarguments, according to \cite{gonzalez2026promoting}, therefore offers significant benefits, pedagogically or otherwise, including

 \begin{enumerate}
 \item It provides evidence of substantive engagement with the source material.
\item It signals that learning has taken place.
\item It develops students' socio-emotional skills as they manage to cultivate an attitude of mutual respect, collaboration and dialogical empathy.
\end{enumerate}

For these reasons, the use of counterargument offers a measurable and theoretically grounded construct for assessing critical thinking in \acrshort{ai}-mediated writing contexts.

\subsection{Thinking Routines}
Thinking routines, which are ways or procedures of thinking systematically, have been shown to develop critical thinking \citep{pinedo2018thinking,manurung2022thinking}.
Many thinking routines exist, e.g. \textit{See-Think-Wonder}, \textit{the 4 C's}, and \textit{circle of viewpoints} \citep{ritchhart2011making}.
Comparing them requires distinguishing between the cognitive processes underlying human reasoning and the procedural mechanisms used to evaluate it.
In the context of argumentative writing, they refer to systematic practices through which individuals construct claims, evaluate evidence, engage opposing views, and justify conclusions.
Examples of thinking routines relevant to argument-based learning (categorized into 3 parts) include \textit{the explanation game} (for introducing and exploring ideas), \textit{connect-extend-challenge} (for synthesizing and organising ideas), and \textit{what makes you say that} ( for digging deeper into ideas) \citep{ritchhart2011making}.
\cite{ritchhart2011making} advocated for flexibility in using the routines, as some examples may cut across multiple categories.
%Educational psychology defines critical thinking as the active, purposeful process of analyzing, evaluating, and constructing reasoning to support claims. .
In the era of \acrshort{gai}, thinking routines may be instantiated differently across human and artificial agents.
While students engage in argumentative writing through cognitive and metacognitive processes, \acrshort{llm}s generate structured arguments through probabilistic pattern recognition trained on large corpora.
%This distinction complicates evaluation: the presence of counterargument structures does not necessarily imply equivalent epistemic engagement. Consequently, comparing thinking routines requires examining both production and evaluation.
%\begin{enumerate}
% \item how students enact argumentative moves
%\item how human evaluators interpret their quality
%\item how AI-based systems operationalize similar constructs
%\end{enumerate}

\subsection{Assessment and Rubrics}

\cite{ku2009assessing} argues that 
simply using multiple-choice response format is inadequate for revealing students’ underlying reasoning for an answer or the ability to think critically under unprompted situations, thereby advocating for assessment that allows both multiple-choice and open-ended format.
Different critical thinking assessment tools exist, e.g. Halpern Critical Thinking Assessment Using Everyday Situations (HCTAES) \citep{halpern2006halpern} and Watson-Glaser Critical Thinking Appraisal (WGCTA) \citep{watson1980watson}, but they differ in their formats and contexts \citep{ku2009assessing}.

Rubrics, as scoring guide for assessing specific components of a task \citep{yavuz2025utilizing,ling2025review}, play a central role in operationalizing abstract constructs such as critical thinking into observable and scorable criteria.
By defining performance dimensions and scale descriptors, rubrics aim to enhance reliability, transparency, and alignment between learning objectives and assessment practices \citep{brookhart2013create, jonsson2007use}. However, the validity of a rubric depends not only on the selected dimensions but also on how clearly performance levels map onto those dimensions.
%\cite{kosmyna2025your} used the following rubrics for grading their essays: Uniqueness, Vocabulary, Grammar, Organization, Content, and Length. In their work, they used a 6-point Likert scale (0: not at all, 1: insufficient, 2: sufficient, 3: satisfactory, 4: good, 5: excellent). It was not all the rubrics that meaningfully matched the Likert scale.
%For example, it is unclear what 'not at all' or 'insufficient' grammar means.
Misalignment raises concerns about construct validity and interpretability, particularly when scales are applied uniformly across heterogeneous rubrics.
Such challenges are especially relevant when comparing human and \acrshort{ai}-based assessments.
%, as \acrshort{llm}s may rely heavily on surface linguistic features (e.g., vocabulary richness or fluency) unless rubric definitions explicitly capture deeper constructs such as argument quality.
Therefore, careful rubric design becomes essential when the goal is to assess counterarguments and critical thinking rather than general writing proficiency.

%Rubric dimensions (e.g.):
%Relevance, Depth, Evidence integration, Refutation strength, Epistemic stance, 

\subsection{\acrshort{ai} Evaluation}

%\subsubsection{LLM-as-a-Judge}
The use of \acrshort{llm}s as evaluators has recently emerged as a promising approach for scalable and cost-efficient assessment.
Rather than serving solely as text generators, frontier models have been shown to approximate human judgments in structured evaluation tasks when provided with explicit criteria and grading rubrics \citep{kocmi2023large,zheng2023judging}.
This paradigm, often referred to as “LLM-as-a-judge,” relies on prompting models to assess outputs according to predefined dimensions, sometimes achieving substantial agreement with expert raters.
Such findings suggest that LLMs can operationalize assessment constructs when evaluation standards are clearly specified.
\acrshort{ai} is particularly useful for scaling assessment when the number of students are so many that it's inconvenient for humans or negatively impacts the quality of assessment humans can do.

Regardless, there are still concerns related to validity, consistency, and sensitivity to deeper cognitive dimensions. For example, \acrshort{gai}-based evaluation can be sensitive to prompt phrasing and rubric formulation, raising questions about robustness and replicability \citep{zheng2023judging}.
It is also well-established that \acrshort{llm}s suffer from hallucinations and other misalignments \citep{adewumi2025ai,adewumi-etal-2025-limitations,adewumi2024fairness}, despite
\acrshort{gai} alignment methods for ensuring that they adhere to human intentions and values \citep{li2026landscape}.
Hence, when evaluating counterarguments, these limitations are especially important.
A text may include a clearly written opposing view, but this does not necessarily mean that the writer has genuinely engaged with the issue or demonstrated deep critical thinking.
For this reason, assessing the performance of an \acrshort{llm}-as-a-judge should go beyond simply comparing scores to examining whether the model’s \acrfull{cot} reasoning is logical or reflects established theoretical definitions of critical thinking.

%This consideration underlies the comparative analysis between human and AI evaluators in the present study.

%\subsubsection{Alignment Theory}

%Interrater reliability: ICC, Cohen’s kappa

% Number of AI runs (stability check)

%  Correlation (Pearson/Spearman), Bias analysis, 

% Statistical Analysis: Multilevel modeling (texts nested within writers), Generalizability theory (if strong journal), Measurement invariance testing, Bland–Altman analysis (for agreement), Regression models predicting divergence

% Does AI overvalue length?
%Regression models predicting discrepancy:

% measurement invariance testing & Report effect sizes and confidence intervals

\section{Materials and Methods}
\label{method}

%\subsection{Research Design}
We combine both qualitative and quantitative research designs for a more insightful intervention study, given that each one presents a unique perspective to the study.
We perform Spearman's correlation analysis for investigating correlation between key rubrics.
We provide additional details of the materials and mixed research design in the following subsections.

\subsection{Participants}
The participants were 36 Master students of the Text Mining course of Luleå University of Technology, Sweden, for the 2025/26 calendar period.\footnote{Course code: D7058E}
The students are nationals of different countries and are mostly around the same age bracket in their early twenties.
The course is based on hybrid onsite-online delivery.
Enrollment in the course meant that students became participants automatically because the counterargument task was designed as one of the assignments of the course.
The students were awarded credits if the task was successfully completed.
Completing the task involved submitting the \acrshort{llm} argument of their choice, their self-written counterargument, and 2 peer-reviews that were randomly assigned on the Canvas \acrfull{lms}.

\subsection{Instruments}

%A thesis statement (or macro-theme) expresses an essay's main idea (Miller and Pessoa, 2016).

Four topics (or thesis statements) of debate were selected after searching online for some modern scientific debates.
The selection was based on the first 4 results from reputable venues across diverse categories, including statistics, linguistics, population genetics, and education.
They include:

\begin{enumerate}
    \item Statistically non-significant results indicate ‘no difference’. \footnote{www.nature.com/articles/d41586-019-00857-9}
    
    \item Humans are born with an innate capacity for language. \footnote{www.simplypsychology.org/naturevsnurture.html}
    
    \item The fate of mutations, which occurs randomly, is singularly governed by natural selection.\footnote{www.tandfonline.com/doi/epdf/10.1080/00219266.2009.9656163? needAccess=true}
    
    \item Pedagogy relates only to ways or methods of teaching.\footnote{www.tandfonline.com/doi/full/10.1111/curi.12006}
    
\end{enumerate}

The students were asked to prompt any \acrshort{llm} of their choice with one of the topics by starting with: \textit{Write an argument for the following thesis statement}.
Thereafter, without any use of \acrshort{gai}, they were required to write their counterargument of 300 words, minimum, to the \acrshort{ai}-generated argument with academic referencing and to submit after 6 days for assessment.

Six established rubrics were employed for assessment, including \textit{Focus}, \textit{Logic}, \textit{Content}, \textit{Style}, \textit{Correctness}, and \textit{References} \citep{howard2012handbook,adewumi2023procot}.
Focus is important because \cite{liu2020counterargumentation} identified in their study on argumentative writing the problem of participants drifting from the topic.
Each was represented on a 5-point Likert scale, which provides ordinal data and is suitable for statistical analysis.
The allocation of the peer-assessment was automatically randomized anonymously on the \textit{Canvas} learning management system and 2 submissions assigned to each student.
This helps to reduce potential bias in grading.
In addition to the Likert scale, there was an optional free-text format for each student to provide open-ended feedback on the rationale for their scores.
The experienced teacher, who is one of the authors, also provided expert assessment for each submission.
The instruction for student pair assessment is given below. 

\begin{quote}

Score each writeup on a Likert scale: 1 (Strongly disagree), 2 (Disagree), 3 (Neither agree or disagree), 4 (Agree), and 5 (Strongly agree) for each of (1) focus, (2) logic, (3) content, (4) style, (5) correctness, and (6) evidence of peer-reviewed references.

\begin{enumerate}
    \item Focus: that the text is focused on the given topic
    \item Logic: that the statements therein are logically constructed
    \item Valid Content: that there is sufficient content for an argument or position
    \item Valid Style: that the style conforms with the academic or prescribed style of writing
    \item Correct: that the argument or position is correct
    \item Peer-Reviewed References: backed up by credible references
    
\end{enumerate}
    
\end{quote}

For cases in the results where averages of the scores are taken, we used the equivalents in Table \ref{likert_av}. 

\begin{table}[h!]
\small
\centering
\caption{Averaging Likert Values}
\label{likert_av}
\begin{tabular}{p{0.3\linewidth} | p{0.3\linewidth}  | p{0.2\linewidth} }
\hline
   \textbf{Likert Scale} & \textbf{Score Range} & \textbf{Equivalent} \\      \hline
 Strongly Disagree & 1.00 - 1.80 & 1 \\
Disagree & 1.81 - 2.60 & 2\\
 Neutral & 2.61 - 3.40 & 3 \\
Agree & 3.41 - 4.20 & 4 \\
Strongly Agree & 4.21 - 5.00 & 5  \\
 \hline
\end{tabular}
\end{table}

\subsection{\acrshort{ai} Models and the System Prompt}

Table \ref{ai_models} identifies the \acrshort{llm}s used in the study.
It includes \acrfull{sota} closed commercial models and open models\footnote{https://huggingface.co/chat/}.
They were implemented through the user interfaces (UIs) and their default hyperparameters were used.
There were two cases (out of 35) that DeepSeek predicted 0, which is not on the Likert scale, for \textit{Reference}.
In both cases, we adjusted the values to the lowest possible values on the Likert scale (1).
The alternative might have been to drop all DeepSeek results completely from calculating the averages but we would be losing valuable information, hence, we settled for adjusting the two anomalies.
The prompt was engineered based on recommendations in the literature for getting the best outputs, including assigning a persona to the model, clarity, and specificity, among others \citep{tripathi2025hitchhiker} and is publicly available.\textsuperscript{1}
As a result, the instruction to the models has slightly more clarification compared to what humans would know by unstated assumption.
The prompt is provided below.

\begin{table}[h!]
\small
\centering
\caption{\acrshort{ai} Models for Counterargument Assessment}
\label{ai_models}
\begin{tabular}{p{0.05\linewidth} | p{0.4\linewidth}  | p{0.4\linewidth} }
\hline
   \textbf{No} & \textbf{\acrshort{llm}} & \textbf{Closed/Open}  \\      \hline
   1 & ChatGPT 5.2 Flagship & Closed \\
2 & ChatGPT 5.2 Instant & Closed \\
   3 & ChatGPT 5.1 Instant & Closed \\
4 & Gemini 3 Thinking & Closed \\
   5 & DeepSeek V3.1 & Open \\
6 & LLaMA 4-Maverick 17B & Open \\
 \hline
\end{tabular}
\end{table}

\begin{quote}
    Assume the role of an expert educator, skilled in assessments. In the zipped file, each document has an argument (or thesis) and counterargument (or response). Score only each counterargument writeup on a Likert scale from 1 (Strongly disagree), 2 (Disagree), 3 (Neither agree or disagree), 4 (Agree), and 5 (Strongly agree) for each part of the rubric below. Optionally, give any comment about the counterargument writeup. Save the complete structured assessment in a downloadable Excel file.

\begin{enumerate}
    \item Focus: that the counterargument is focused on the given topic (by comparing the counterargument with the original argument)
    
    \item Logic: that the statements in the counterargument are logically constructed
    
    \item  Valid Content: that there is sufficient content (minimum of 300 words) for the counterargument
    
    \item Valid Style: that the style conforms with the academic style of writing
    
    \item Correct: that the counterargument is correct 
    
    \item Peer-Reviewed References: that the counterargument is backed up by credible references
\end{enumerate}

\end{quote}

For Gemini and the open models, the clause \textit{In the zipped file} was omitted and the following alternative sentence substituted the last sentence because they were not able to generate Excel file as the final output - \textit{Write out the results in a tabular form showing the provided filenames and the scores per file for all rubrics}.

\subsection{\acrshort{ai} Checks of Counterarguments}
We used two popular \acrshort{ai}-text-generation checkers on the students' counterarguments: Grammarly's \acrshort{ai} detector and ZeroGPT.\footnote{grammarly.com/ai-detector and zerogpt.com}
We realized from the evaluations that these systems are not that reliable or consistent for checking \acrshort{ai}-generated text (see Table \ref{ai_checks} in the appendix).
For example, comparing each output in the two systems, the average, maximum, and minimum differences are 25.98\%, 79.8\% (or standard deviation of 39.9\%), and 0, respectively.
Furthermore, 16 cases have differences above 20\%.
As a result, we relied on the expert's experience to determine which of the submissions was largely (or wholely) \acrshort{ai}-generated, which happened to be one out of the 36 submissions.
We confirmed this from the student concerned and they admitted it to be so.
Hence, we left the submission out of this study.

\subsection{Ethics Consideration}

The students peer-assessment was anonymized to reduce the possibility of bias in assessment among students.
We also ensured that we checked each counterargument submission for \acrshort{ai}-generated content.

\section{Results}
\label{results}

\begin{figure}[h!]
\centering
\includegraphics[width=0.5\textwidth]{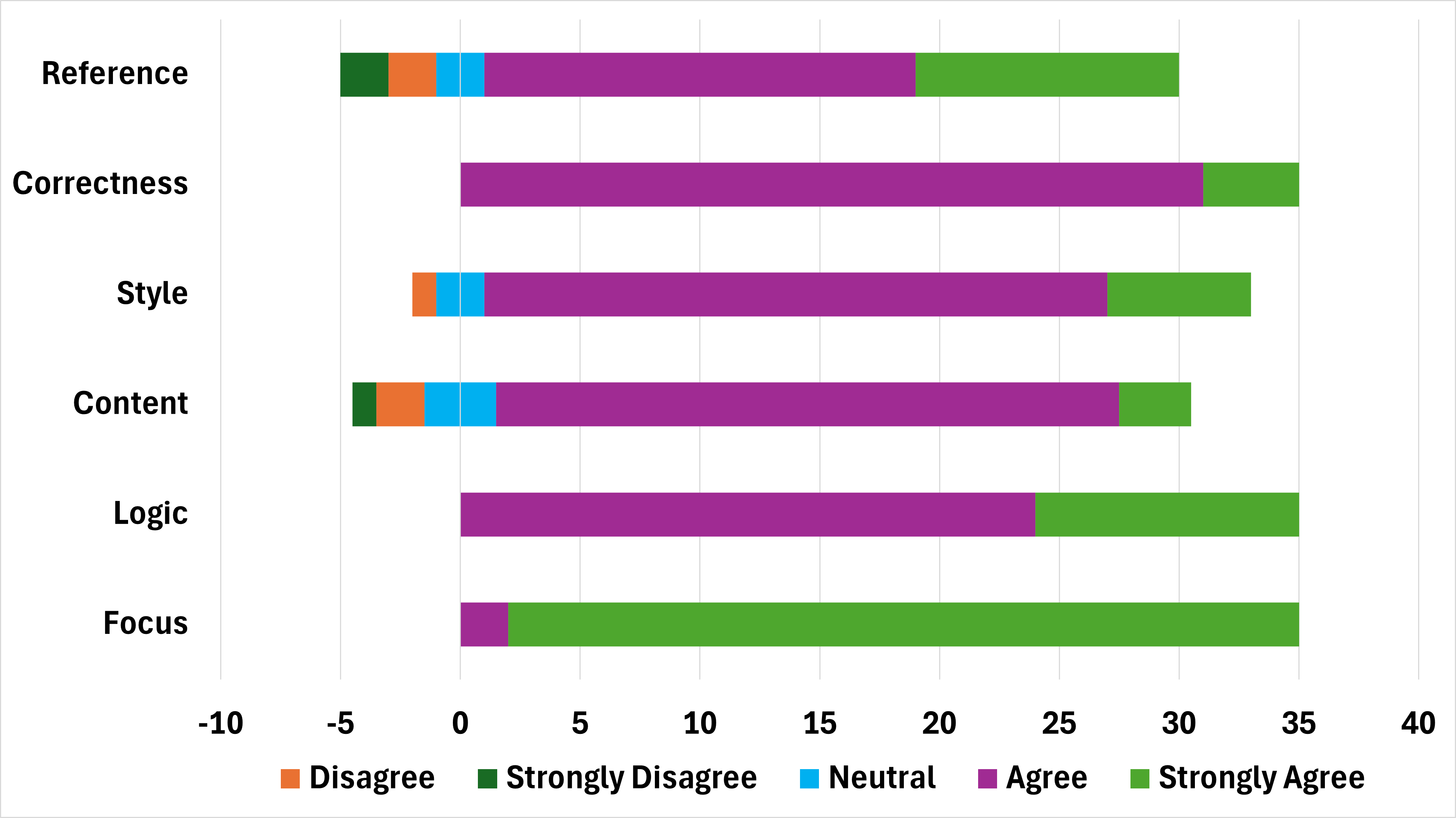}
\caption{Diverging Stacked bar chart for Expert Assessment.} 
\label{fig_divergingbar}
\end{figure}

\begin{figure}[h!]
\centering
\includegraphics[width=0.5\textwidth]{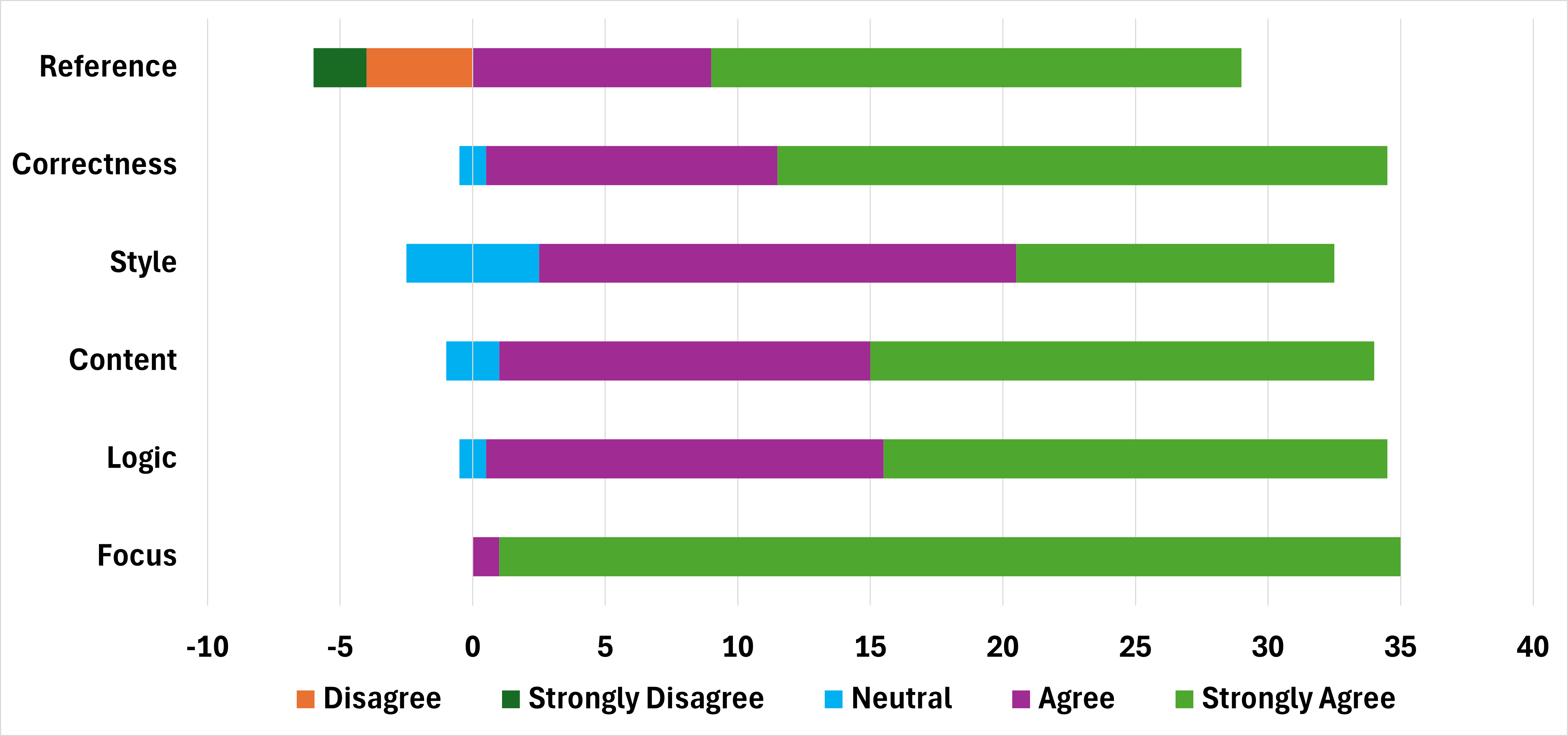}
\caption{Diverging Stacked bar chart for Average Student Assessment.} 
\label{fig_divergingbar_pair}
\end{figure}

\begin{figure}[h!]
\centering
\includegraphics[width=0.5\textwidth]{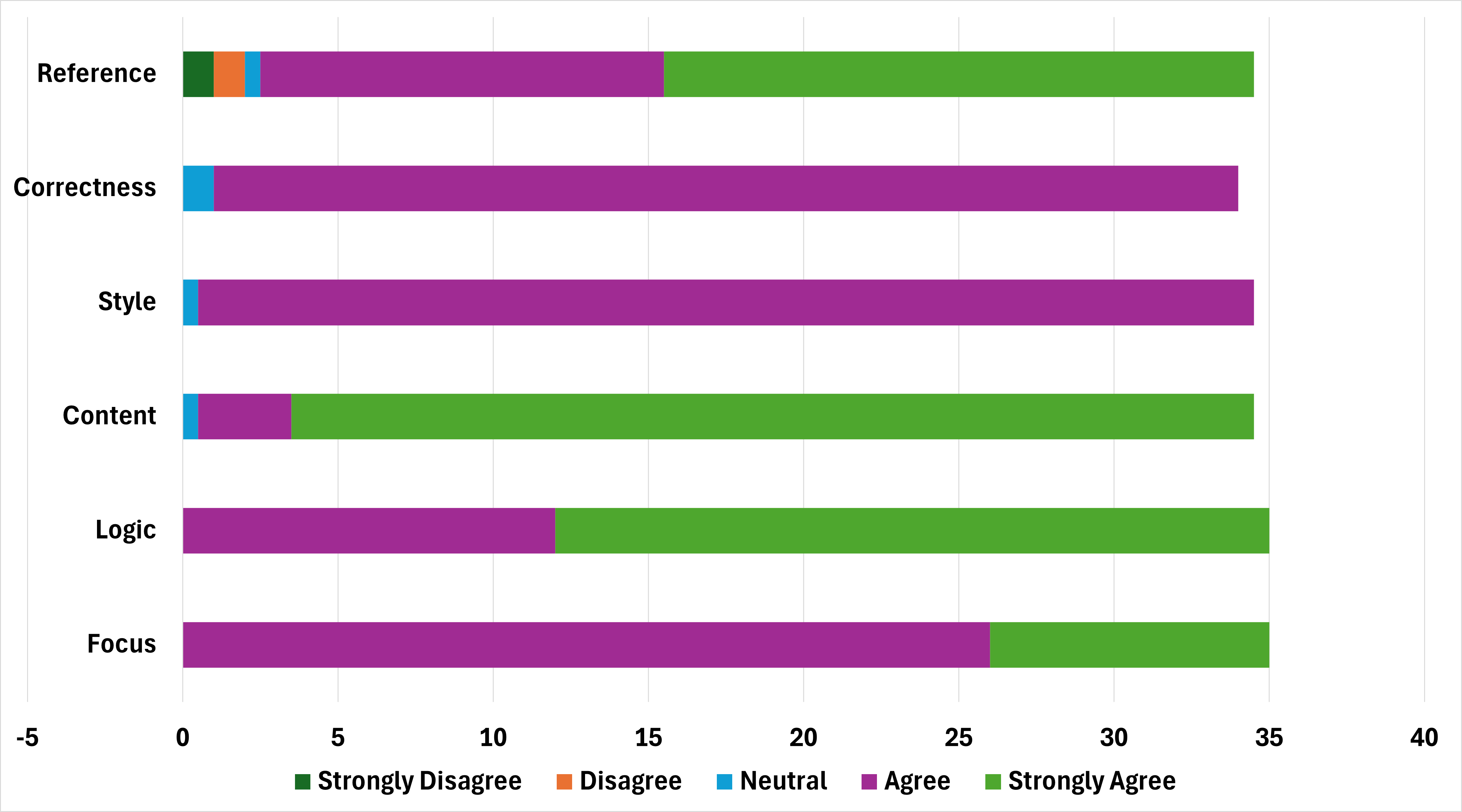}
\caption{Diverging Stacked bar chart for Average \acrshort{ai} Assessment.} 
\label{fig_divergingbar_ai}
\end{figure}

We present the results of the study from multiple perspectives.
For \acrshort{ai}-generated arguments by students, about 82\% of the \acrshort{ai} models used was \textit{ChatGPT 5.1} while 12\% and 6\% were \textit{Copilot} and \textit{ChatGPT 5.2 Thinking}, respectivley.
Figures \ref{fig_divergingbar}, \ref{fig_divergingbar_pair}, and \ref{fig_divergingbar_ai} represent diverging stacked bar charts for the expert assessment, average student assessment, and the average \acrshort{ai} assessment of the counterarguments, respectively.
We can observe that all the three charts show that the rubrics for most counterarguments appear to the right of the scale, being \textit{Agree} or \textit{Strongly Agree}.
There are strong similarities in the rubrics for the expert and the average student assessments, as well as the \acrshort{ai} assessment, in some cases while they are not as strong in some others.
The most important rubric, \textit{Logic}, shows that the expert appears stricter in assessment (with only 10 as \textit{Strongly Agree}) than both the average student and average \acrshort{ai}.

%For example, the length of the bar with \textit{Strongly Agree} for \textit{Focus} is very similar for both while .

%Best for: Showing the full distribution of responses while highlighting agreement vs. disagreement.

\begin{figure}[h!]
\centering
\includegraphics[width=0.5\textwidth]{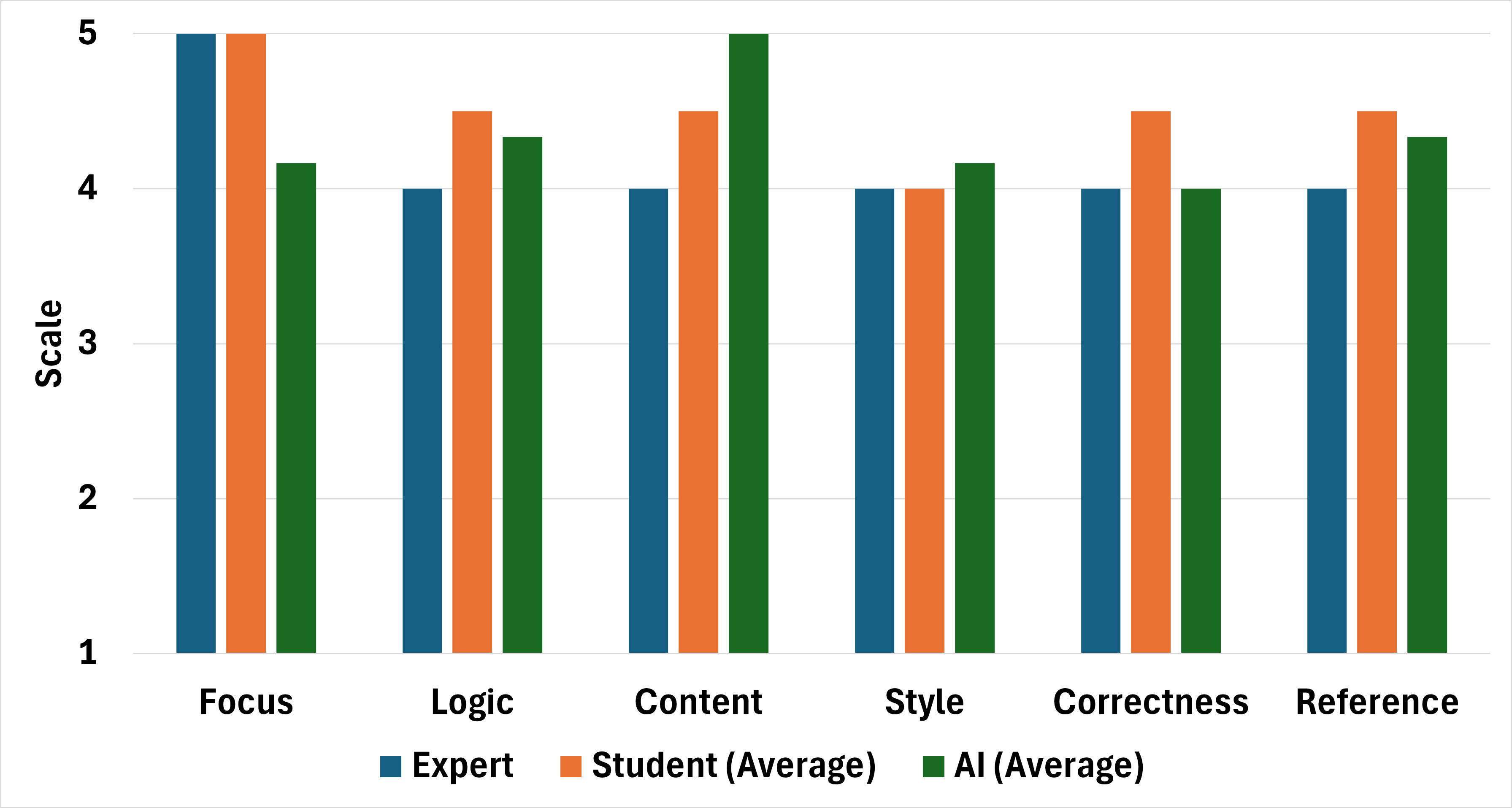}
\caption{Bar chart of medians.} 
\label{fig_medianbar}
\end{figure}

\begin{figure}[h!]
\centering
\includegraphics[width=0.5\textwidth]{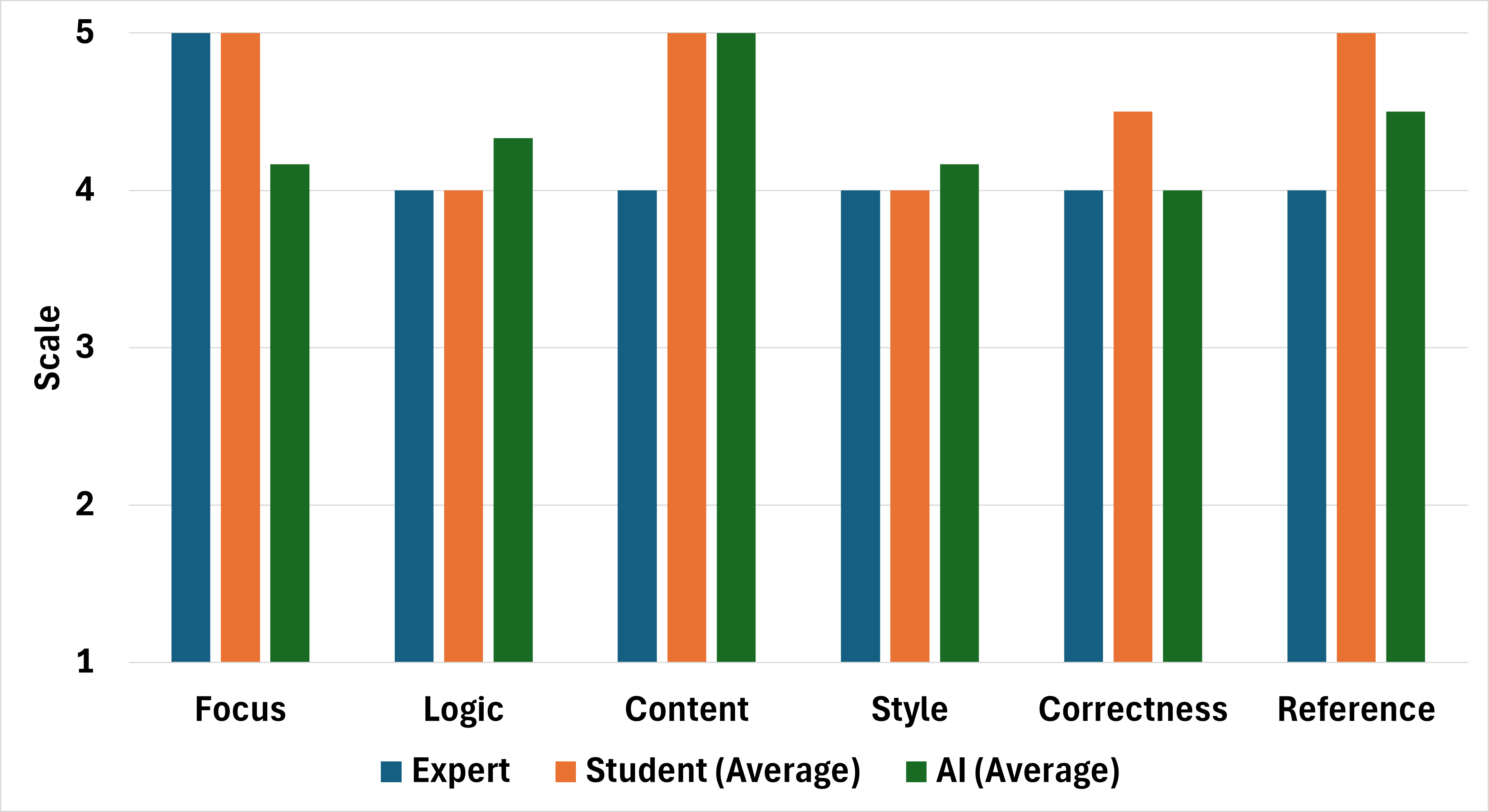}
\caption{Bar chart of modes.} 
\label{fig_modebar}
\end{figure}

Figures \ref{fig_medianbar} and \ref{fig_modebar} represent the bar charts of medians and modes, respectively, for the 3 types of assessments: expert, average student, and average \acrshort{ai}.
The median values for all rubrics are 4 and above, indicating \textit{Agree} and \textit{Strongly Agree} while the modes are equally 4 and above, also indicating \textit{Agree} and \textit{Strongly Agree}.

\begin{figure}[h!]
\centering
\includegraphics[width=0.5\textwidth]{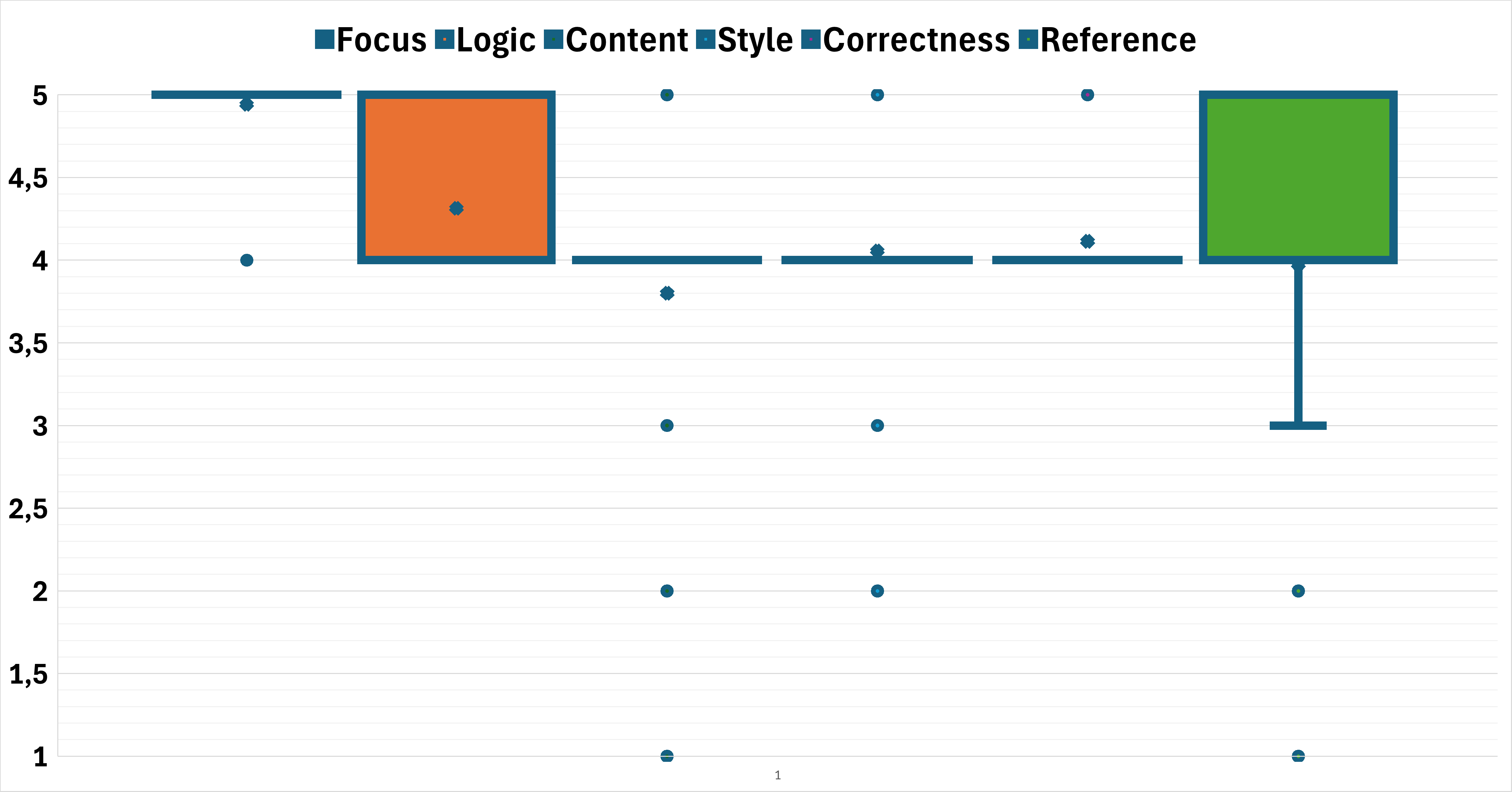}
\caption{Box Plot of Expert Assessment.} 
\label{fig_box1}
\end{figure}

\begin{figure}[h!]
\centering
\includegraphics[width=0.5\textwidth]{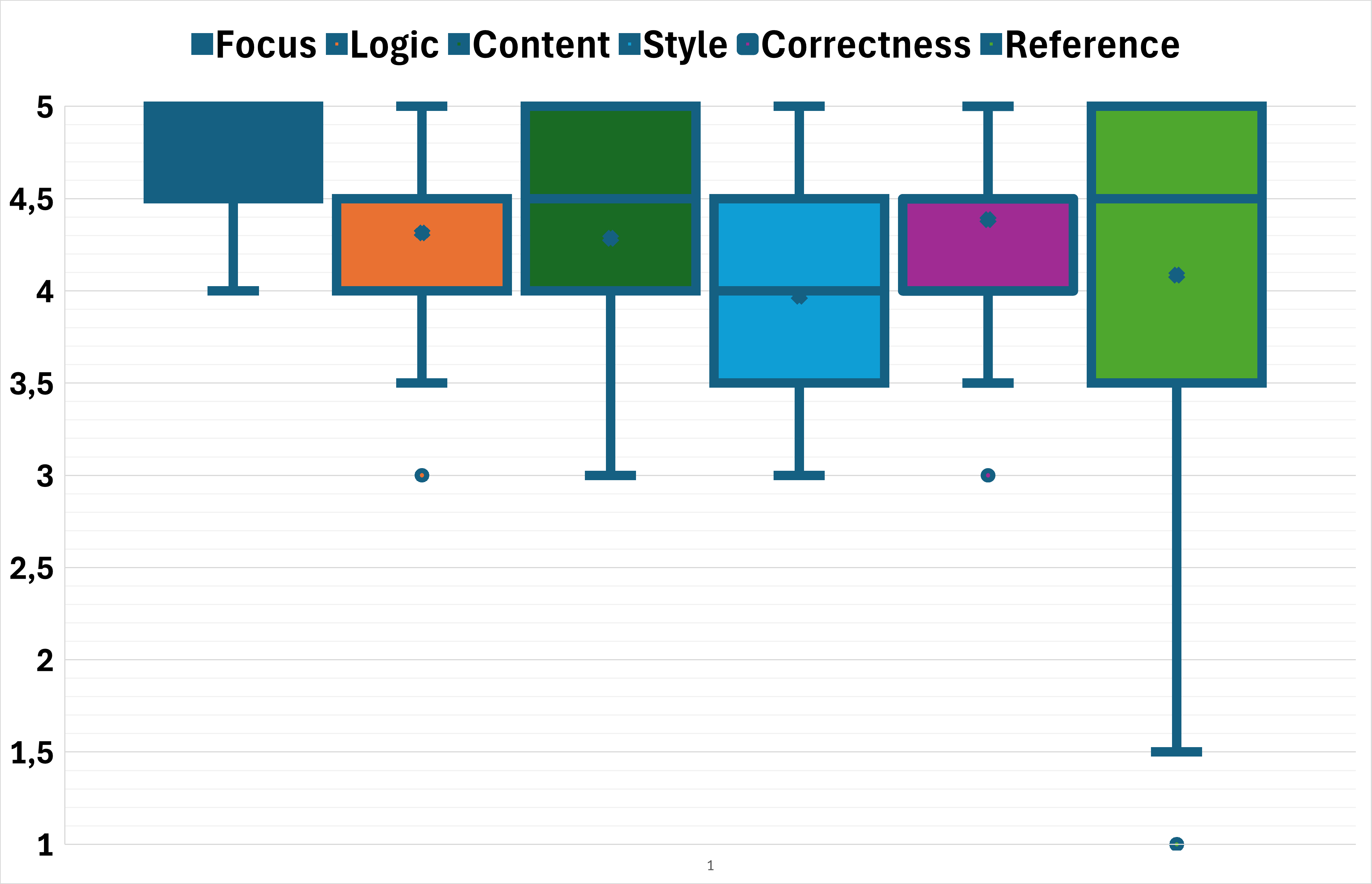}
\caption{Box Plot of Average Pair Assessment.} 
\label{fig_box2}
\end{figure}

\begin{figure}[h!]
\centering
\includegraphics[width=0.5\textwidth]{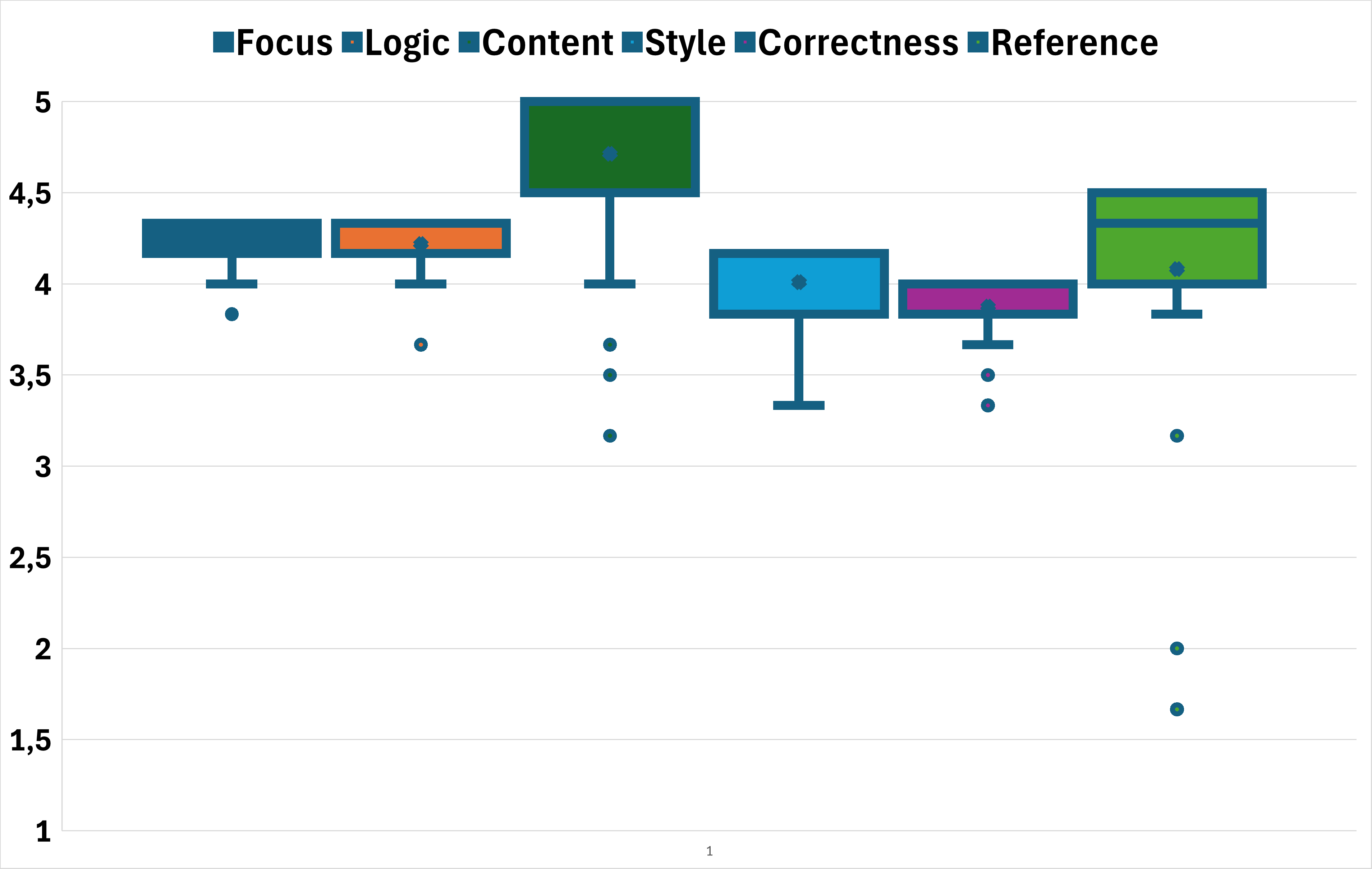}
\caption{Box Plot of Average \acrshort{ai} Assessment.} 
\label{fig_box3}
\end{figure}

Figures \ref{fig_box1}, \ref{fig_box2}, and \ref{fig_box3} represent the box and whisker plots for the expert, average student, and the average \acrshort{ai}, respectively.
The three box plots show more peculiarities per assessment or average assessment.
For example, the plot for expert assessment shows 4 of the rubrics have collapsed boxes, indicating the dominant value, and several outliers.
The average student assessment has the most expanded boxes and fewer outliers compared to the other two.
The plot for the average \acrshort{ai} assessment has narrow boxes and similar count of outliers as the plot for expert assessment.

\subsection*{Additional Analysis}

Tables \ref{correlat} and \ref{sub_correlat} present average values of all rubrics for all submissions and for the 2 highest categories of submissions (linguistics and education), respectively.
The Gwets AC2 inter-rater reliability values for all the models are 0.33, except 0.17 for \acrshort{llama}, which are all lower than students' agreement values with the expert.
We used Gwet's AC2 because it handles concerns about the paradoxes of Cohen's Kappa for measuring inter-rater agreement.
Furthermore, using the average values from Table \ref{correlat} to perform Spearman's correlation analysis of key rubrics, we observe a very strong positive monotonic correlation between logic and focus (0.849) and logic and correctness (0.919).
A similar observation is apparent when we consider only the six \acrshort{llm}s.
We observe a very strong positive monotonic correlation between logic and focus (0.882) and logic and correctness (0.893).
The table also shows the 2 ChatGPT 5.2 models have the largest average difference between the two correlated rubrics \textit{Logic} and \textit{Correctness}, though one would have expected a smaller difference, as with others.
ChatGPT 5.1 deviates most from others (with the average 3 \textit{Neither agree or disagree}) on the most important rubrics.

\begin{table*}[h!]
\small
\centering
\caption{Average Values of All Rubrics Across \acrshort{ai} and Humans}
\label{correlat}
\begin{tabular}{p{0.02\linewidth} | p{0.18\linewidth}  | p{0.05\linewidth} | p{0.05\linewidth} | p{0.07\linewidth} | p{0.05\linewidth} | p{0.09\linewidth} | p{0.07\linewidth} | p{0.04\linewidth} }
\hline
   \textbf{No} & \textbf{Entity} & \textbf{Focus} & \textbf{Logic} & \textbf{Content} & \textbf{Style} & \textbf{Correctness} & \textbf{Reference}  & \textbf{Gwet's AC2}  \\      \hline
   1 & ChatGPT 5.2 Flagship & 3.314
 & 4 & 5 & 3 & 3 & 4.229 & 0.33\\
2 & ChatGPT 5.2 Instant & 4 & 4 & 4,829
 & 4 & 3 &  2.943 & 0.33\\
   3 & ChatGPT 5.1 Instant & 3 & 3 & 4.429 & 3 & 3 & 3.829  & 0.33\\
4 & Gemini 3 Thinking & 5 & 4.8 & 4.2 & 4.629 & 4.829 & 4.743 & 0.33\\
   5 & DeepSeek V3.1 & 4.886 & 4,629  & 4.829 & 4.6 & 4.657 & 4.257  & 0.33\\
6 & \acrshort{llama} 4-Maverick 17B & 4.886 & 4.886 & 5 & 4.829 & 4.771 & 4,486 & 0.17 \\
7 & Expert (Teacher) & 4.943 & 4.314 & 3.8 & 4.057 & 4.114 & 3.971 & 1 \\
8 & Student 1 & 4.8 & 4.342 & 4.286 & 3.914 & 4.257 & 4.086 & 1\\
9 & Student 2 & 4.743 & 4.286 & 4.286 & 4.029 & 4.514 & 4.086 & 0.83\\
 \hline
\end{tabular}
\end{table*}

\begin{table*}[h!]
\small
\centering
\caption{Average Values of All Rubrics for 2 Topic Across \acrshort{ai} and Humans}
\label{sub_correlat}
\begin{tabular}{p{0.05\linewidth} | p{0.2\linewidth}  | p{0.05\linewidth} | p{0.05\linewidth} | p{0.07\linewidth} | p{0.05\linewidth} | p{0.09\linewidth} | p{0.07\linewidth} }
\hline
   \textbf{No} & \textbf{Entity} & \textbf{Focus} & \textbf{Logic} & \textbf{Content} & \textbf{Style} & \textbf{Correctness} & \textbf{Reference}  \\      \hline
\multicolumn{8}{l}{Linguistics (\textit{n = 14})}  \\
   1 & ChatGPT 5.2 Flagship & 3.643 & 4 & 5 & 3 & 3 & 4.143 \\
2 & ChatGPT 5.2 Instant & 4 & 4 & 5 & 4 & 3 & 2.929 \\
   3 & ChatGPT 5.1 Instant & 3 & 3 & 4.714 & 3 & 3 & 3.786  \\
4 & Gemini 3 Thinking & 5 & 4.786 & 4 & 4.5 & 4.857 & 4.714 \\
   5 & DeepSeek V3.1 & 4.857 & 4.571 & 4.786 & 4.5 & 4.571 & 4.071 \\
6 & \acrshort{llama} 4-Maverick 17B & 4.929 & 4.929 & 5 & 4.786 & 4.643 & 4.5 \\
7 & Expert (Teacher) & 5 & 4.357 & 3.929 & 4.214 & 4.214 & 3.857 \\
8 & Student 1 & 4.571 & 4.143 & 4.143 & 3.929 & 4.286 & 3.714 \\
9 & Student 2 & 4.786 & 4.286 & 4.429 & 3.929 & 4.286 & 3.857 \\
 \hline

 \multicolumn{8}{l}{Education (\textit{n = 13})}  \\
   1 & ChatGPT 5.2 Flagship & 3.154 & 4 & 5 & 3 & 3 & 4.308 \\
2 & ChatGPT 5.2 Instant & 4 & 4 & 4.769 & 4 & 3 & 2.923
 \\
   3 & ChatGPT 5.1 Instant & 3 & 3 & 4.077 & 3 & 3 & 3.769 \\
4 & Gemini 3 Thinking & 5 & 4.692 & 4.231 & 4.615 & 4.692 & 4.615 \\
   5 & DeepSeek V3.1 & 4.846 & 4.462 & 4.769 & 4.462 & 4.538 & 4.077
 \\
6 & \acrshort{llama} 4-Maverick 17B & 4.769 & 4.769 & 5 & 4.769 & 4.769 & 4.154 \\
7 & Expert (Teacher) & 4.846 & 4.231 & 3.538 & 4 & 4 & 3.846 \\
8 & Student 1 & 4.769 & 4.385 & 4.385 & 4.154 & 4.692 & 4.308 \\
9 & Student 2 & 4.692 & 4.231 & 4 & 3.769 & 4 & 3.923 \\
 \hline

\end{tabular}
\end{table*}

\begin{figure}[h!]
\centering
\includegraphics[width=0.5\textwidth]{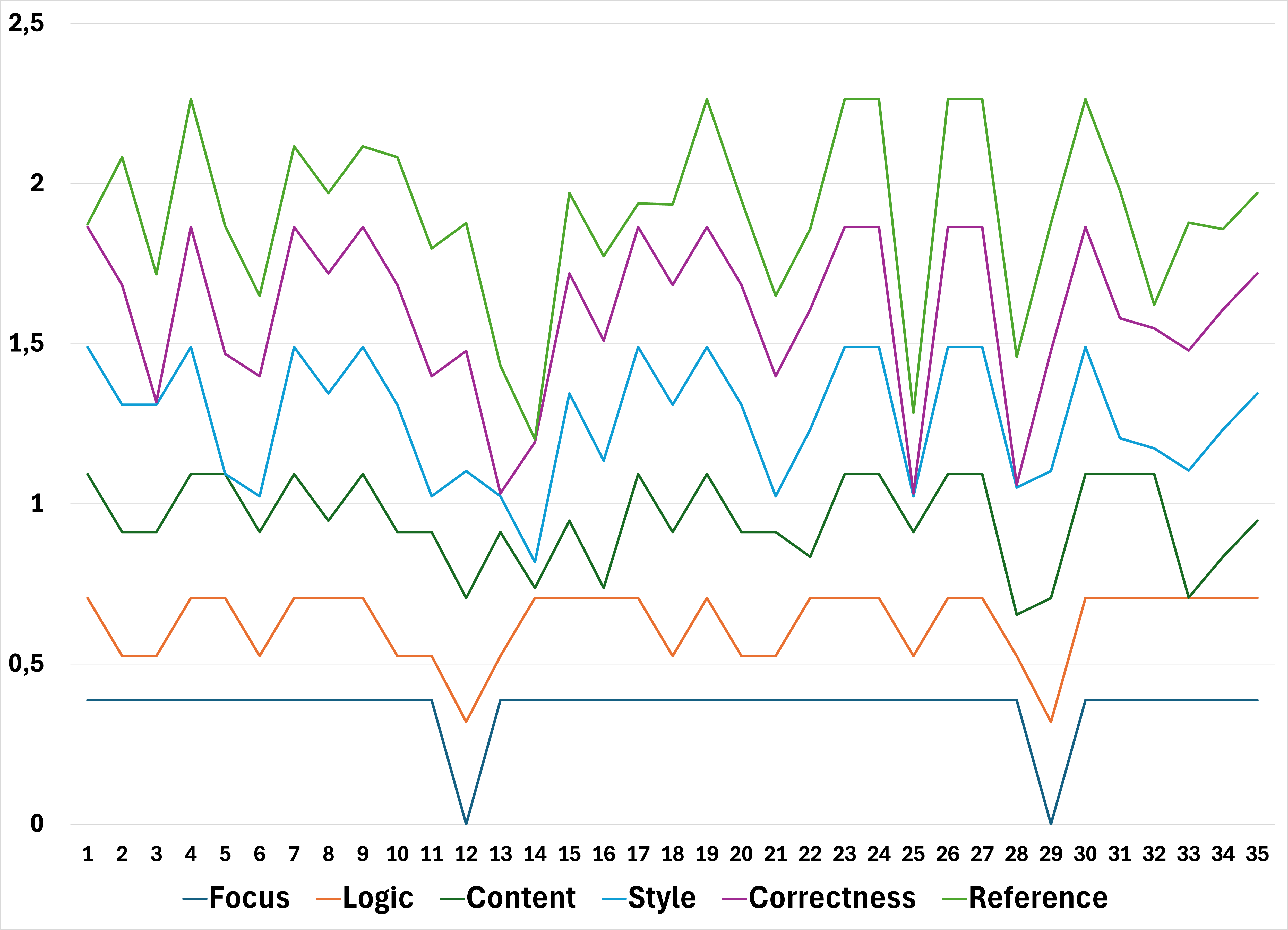}
\caption{Z score distribution for expert reviews.} 
\label{fig_zscore}
\end{figure}

Figure \ref{fig_zscore} represents the Z-score distribution chart for the expert assessment for all the 35 counterarguments.
The chart helps to determine how many standard deviations each value for all the rubrics is from the average.
We can observe from the chart that \textit{Focus}, followed by \textit{Logic}, is the rubric with the least distance from the mean while \textit{Reference} is the one with the most (with some values above 2), though not extreme.

%\subsection{Qualitative Results}
%When we consider some individual cases, as identified in Table \ref{prediction_cot}.

\begin{table*}[h!]
\small
\centering
\caption{Reasoning examples for a Couple of Submissions}
\label{prediction_cot}
\begin{tabular}{p{0.01\linewidth} | p{0.26\linewidth}  | p{0.67\linewidth} }
\hline
   \textbf{\#} & \textbf{Model (Likert Scale: F L C S C R)} & \textbf{Reasoning (optional)}  \\
\hline
1 & ChatGPT 5.2 Instant (4 4 5 4 3 2) & Because the rubric’s criteria (logic, correctness, focus, style) require human judgment, the automated model used reasonable heuristics: Valid Content: based on word count (>300 words = strong). Peer-Reviewed References: checked for publication years or links. Other categories neutral baseline values (3), since deeper qualitative judgment requires human reading unless we define more explicit rules. If you want more nuanced scoring, e.g., lexical cohesion checks, citation detection, topic alignment modeling—I can upgrade the rubric engine. \\
& & \\
 & Gemini 3 Thinking (5 4 2 3 4 1)
 & Good logical flow regarding "pattern-finding" , but lacks any academic citations and falls short on word count. \\
 &  &  \\
 & DeepSeek V3.1 (5 4 5 4 4 0) & A well-structured, focused, and substantive counterargument that effectively challenges the nativist position. It clearly articulates the role of social interaction and pattern recognition. The major flaw is the complete lack of citations for its claims, which severely undermines its academic credibility. \\
   & & \\
 & LLaMA 4-Maverick 17B (5 5 5 5 4 4) & \\
& & \\
 & Expert (5 4 4 4 4 1) & The focus of the counterargument is strongly on the topic and \acrshort{ai}-generated argument, as every paragraph discusses the issue.
 The logic is strong and uses examples to make the case.
 The reference to "too little" in the counterargument has no basis in the original argument - though this may have been referring to the word "incomplete," which are not necessarily the same. The word count is above the minimum, at 317.
 There were 5 paragraphs but one had only 2 sentences.
 While some of the assertions are correct, there was no single reference to the literature though it speaks of studies. This greatly affects the credibility of the claims. \\
& & \\
 & Student 1 (5 4 4 3 4 1) &  \\
 & & \\
 & Student 2 (4 4 4 3 4 2) &  \\
 & & \\
 \hline
 
2 & ChatGPT 5.2 Instant (4 4 5 4 3 3) & Because the rubric’s criteria (logic, correctness, focus, style) require human judgment, the automated model used reasonable heuristics: Valid Content: based on word count (>300 words = strong). Peer-Reviewed References: checked for publication years or links. Other categories neutral baseline values (3), since deeper qualitative judgment requires human reading unless we define more explicit rules. If you want more nuanced scoring, e.g., lexical cohesion checks, citation detection, topic alignment modeling—I can upgrade the rubric engine. \\
& & \\
 & Gemini 3 Thinking (5 5 5 5 5 5)
 & Very thorough. Successfully integrates philosophy, leadership, and social equity to broaden the definition. \\
 &  &  \\
 & DeepSeek V3.1 (5 5 5 5 5 5) & A superb counterargument. It is highly focused and systematically dismantles the narrow definition of pedagogy by introducing equity, diversity, leadership, and philosophy. The logic is clear and the writing is academic. It is backed by four credible, peer-reviewed references, making it a very strong piece of work. \\
   & & \\
 & \acrshort{llama} 4-Maverick 17B (5 5 5 5 5 5) & \\
& & \\
 & Expert (5 5 4 5 4 5) & The focus of the counterargument was consistently on the topic and original argument.
 The logic included clear definitions, historical viewpoints, and why the original argument is not good enough to stand.
 The 5 paragraphs were substantial in content though some grammar issues were observed.
 Bold text was used in some cases for emphasizing points. 
 The word count is 483, offering much to say with correct points that are supported by 4  relevant references, which are used in-text.\\
& & \\
 & Student 1 (4 4 4 4 4 4) &  \\
 & & \\
 & Student 2 (5 5 5 5 4 5) &  \\
 \hline
\end{tabular}
\end{table*}

%\subsection{Counterarguments by \acrshort{llm}s}
\section{Discussion}
\label{discussion}

%\subsection{Challenges with human brain when using AI}

While recent advances in \acrshort{gai} have begun to demonstrate significant potential in education for autonomous learning \citep{liu_role_2025}, enhancing student engagement \citep{rahman_chatgpt_2023}, and supporting writing and idea generation \citep{kasneci2023chatgpt}, %Moreover, the dual role of AI-driven tools can enhance learning outcomes by offering adaptive and immediate feedback, which supports the acquisition of skills and knowledge retention \citep{kwak_effectiveness_2025}. However, alongside these benefits,
emerging research has highlighted potential negative effects on cognitive functions, including perception, learning, critical thinking, problem solving, and decision-making abilities \citep{gerlich_ai_2025}. 
Growing concerns have suggested that extensive reliance on \acrshort{gai} tools in academic tasks may have implications for cognitive development, particularly with respect to independent problem solving and critical thinking \citep{kosmyna_your_2025}.
The results of this study indicate that argument-based learning with \acrshort{gai} promotes critical thinking because of the strong agreement that statements in the counterarguments are logically constructed, generally, and focused on the given topic.
Building on Kuhn’s framework, comparing thinking routines in argument-based learning or any type of learning, involves examining how argumentative skills are enacted and recognized across different evaluative agents. In human reasoning, the production of counterarguments and rebuttals reflects deliberate epistemic coordination and dialogic engagement \citep{kuhn2018role}, consistent with models of argument structure that emphasize rebuttal as a marker of sophisticated reasoning \citep{toulmin2003uses}. These theoretical models position counterargument and rebuttal not as optional rhetorical devices but as core indicators of epistemic maturity.

%\cite{kuhn1991skills} defined a framework of critical thinking whose key aspects include distinguishing claims from evidence, generating and fairly representing counterarguments, constructing rebuttals, and evaluating the relevance and sufficiency of evidence.  He also emphasizes epistemic development, particularly the capacity to recognize that knowledge claims require justification and must be weighed against competing alternatives. Together, these skills reflect the shift from opinion-based reasoning to reflective, evidence-based critical thinking. 

%With constant interaction with AI during argumentative writing, core components of cognitive effort may be externally delegated. This phenomenon has been described as cognitive offloading \citep{risko_cognitive_2016}, and involves transferring cognitively demanding processes to external technological tools in order to alleviate mental burdens. While such offloading can increase efficiency, sustained dependence can be detrimental to the development of active cognitive engagement and reduce opportunities for skill development \citep{zhang_you_2024}. Related findings by \citet{sparrow_google_2011} indicated that the ready availability of external information can influence memory processes and reduce deeper cognitive processing. 

The transition from active information seeking to consumption of structured AI-generated content has further reshaped how learners engage with new knowledge and reasoning process. Unlike conventional search engines, which provide users with multiple sources requiring comparison and evaluation, \acrshort{gai} typically delivers a single synthesized response.
While such responses may discourage cognitive engagement to analyse and evaluate information critically \citep{kasneci2023chatgpt}, they appear to challenge the construction of logical response in argument-based learning.
%Overall, these concerns highlight 
Despite this study, there is the need for continued research on the interaction between \acrshort{gai} and the human brain in other learning paradigms, especially regarding the influence on cognitive skill development.
It is imperative to promote cognitive engagement across diverse tasks while ensuring that \acrshort{gai} tools are used to support, rather than replace, core thinking processes.

%\subsection{Engagement and learning through counterarguments}

\section{Conclusion}
\label{conclusion}

The ongoing transformation in education by \acrshort{gai} looks set to continue.
It appears every area will be affected and learning needs to adapt to this changes.
In this work, we showed that, in argument-based learning, students’ self-written counterarguments to
\acrshort{ai}-generated content promotes critical thinking because they contain logic, in addition to other important components.
Furthermore, we showed that \acrshort{gai} can be successfully used to assess
students’ counterarguments
based on clear rubrics and these assessments generally align
with expert and students' assessments.
Future work needs to evaluate the impact \acrshort{gai} has on other types of learning and other areas of education.

%\section*{Limitations}
% Single cultural context?Prompt sensitivity? Rubric dependency?

\section*{Acknowledgments}

The authors wish to thank the Department of Computer 
Science, Electrical \& Space Engineering at Luleå University of Technology for the 2026 SRT pedagogy fund for this project.
We also thank all the participating students of the Text Mining course, 2025/26 session.
This work was partially supported by the Wallenberg AI, Autonomous Systems and Software Program (WASP), 
funded by the Knut and Alice Wallenberg Foundation and counterpart funding from Luleå University of Technology (LTU).

%\begin{table*}[h!]
%\small
%\centering
%\label{sota_results}
%\begin{tabular}{p{0.02\linewidth} | p{0.1\linewidth} | p{0.16\linewidth} | p{0.16\linewidth} | p{0.16\linewidth} | p{0.16\linewidth}}
%\hline
%   \textbf{No} & \textbf{Judge}   & \textbf{Topic 1 Counterargument} & \textbf{Topic 2 Counterargument} & \textbf{Topic 3 Counterargument} & \textbf{Topic 4 Counterargument} \\      \hline
%   & \textbf{\acrshort{ai}} & & & &  \\
%   1 & ChatGPT 5.2 Instant &  &  & &  \\

% \hline
     %&  Totals & &  &\\ \hline \hline
% \end{tabular}
%\caption{\acrshort{llm}-as-a-Judge and Human evaluations}
% \end{table*}

\appendix

\section{Appendix}
\label{appendix}

\begin{table}[h!]
\tiny
\centering
\caption{Checks for \acrshort{ai}-generated Counterarguments}
\label{ai_checks}
\begin{tabular}{p{0.02\linewidth} | p{0.15\linewidth} | p{0.13\linewidth}  | p{0.13\linewidth} | p{0.12\linewidth}}
\hline
\textbf{No} & \textbf{Grammarly \acrshort{ai} Detector} & \textbf{ZeroGPT} & \textbf{Difference} & \textbf{St. Deviation} \\
\hline
 
1 & 0 & 38,7 & 38,7	& 19.35 \\
2 & 58 & 43,5 & 14,5 & 7.25 \\
3 & 10 & 15,5 & 5,5	& 2.75 \\
4 & 30 & 79,6 & 49,6 & 24.8 \\
5 & 46 & 39	& 7 & 3.5 \\
6 & 0 & 7,3 & 7,3 & 3.65 \\
7 & 8 & 27,6 & 19,6 & 13.8 \\
8 & 0 & 12,9 & 12,9 & 6.45 \\
9 & 0 & 12,9 & 12,9 & 6.45 \\
10 & 0 & 2,4 & 2,4 & 1.2 \\
11 & 30& 79,6 & 49,6 & 24.8 \\
12 & 0 & 9,3	& 9,3	& 4.65 \\
13 & 59	& 98,8 & 39,8 & 19.9 \\
14 & 17 & 92,8 & 75,8 & 37.9 \\
15 & 0 & 26,54 & 26,54 & 13.27 \\
16 & 0 & 34,47 & 34,47 & 17.235 \\
17 & 16 & 17,3 & 1,3	& 0.65 \\
18 & 61 & 70,5 & 9,5 & 4.75 \\
19 & 40 & 28,9 & 11,1 & 5.55 \\
20 & 0 & 40,6 & 40,6 & 20.3 \\
21 & 55 & 48,9 & 6,1	& 3.05 \\
22 & 0 & 18,2 & 18,2 & 9.1 \\
23 & 9 & 46,5 & 37,5 & 18.75 \\
24 & 0 & 18,2 & 18,2	& 9.1 \\
25 & 51 & 75,8 & 24,8 & 12.4 \\
26 & 84 & 92,8 & 8,8	& 4.4 \\
27 & 0 & 0 & 0 & 0 \\
28 & 51 & 29	& 22 & 11 \\
29 & 39 & 23,1 & 15,9 & 7.95 \\
30 & 78 & 23,1 & 54,9 & 27.45 \\ 
31 & 24 & 85,9 & 61,9 & 30.95 \\
32 & 81 & 100 & 19 & 9.5 \\
33 & 0 & 49,3 & 49,3 & 24.65 \\
34 & 9 & 88,8 & 79,8	& 39.9 \\
35 & 68 & 92,6 & 24,6 & 12.3 \\

 \hline
\end{tabular}
\end{table}

%\section{Appendix title 2}
%% \label{}

%% If you have bibdatabase file and want bibtex to generate the
%% bibitems, please use
%%
\bibliographystyle{elsarticle-harv} 
\bibliography{example}

%% else use the following coding to input the bibitems directly in the
%% TeX file.

%%\begin{thebibliography}{00}

%% \bibitem[Author(year)]{label}
%% For example:

%% \bibitem[Aladro et al.(2015)]{Aladro15} Aladro, R., Martín, S., Riquelme, D., et al. 2015, \aas, 579, A101

%%\end{thebibliography}

\end{document}